\documentclass[11pt,a4paper,reqno]{amsart}
\usepackage{multirow}
\usepackage[centertags]{amsmath}
\usepackage{amsfonts}
\usepackage{amssymb}
\usepackage{amsthm}
\usepackage{newlfont}
\usepackage{fullpage,a4}
\usepackage[pdftex]{graphicx,color}

\theoremstyle{definition}
\newtheorem{defn}{Definition}[section]
\newtheorem{theorem}[defn]{Theorem}

\theoremstyle{remark}

\usepackage{bbm}

\def\R{\mathbb R}
\def\Z{\mathbb Z}

\def\r{\rangle}
\def\l{\langle}
\def\rb{\right\rbrace}
\def\lb{\left\lbrace}
\def\affW{W^{\mathrm{aff}}}

\def\a{\alpha}
\def\o{\omega}
\def\ov{\omega^\vee}
\newcommand{\ep}{\varepsilon}
\newcommand{\la}{\lambda}

\def\stab{\operatorname{stab}}
\def\inter{\operatorname{int}}

\newcommand{\ldef}{\mathrel{\mathop:}=}
\newcommand{\sca}[2]{\langle #1,\, #2\rangle}
\newcommand{\set}[2]{\left\{#1 \, |\, #2 \right\}}

\begin{document}

\title{Weyl group orbit functions in image processing}

\author[G. Chadzitaskos]{Goce Chadzitaskos}
\author[L. H\'akov\'a]{Lenka H\'akov\'a}
\author[O. Kaj\'inek]{Ond\v{r}ej Kaj\'inek}

\date{\today}

\begin{abstract}
We deal with the Fourier-like analysis of functions on discrete grids in two-dimensional simplexes using $C-$ and $E-$ Weyl group orbit functions. For these cases we present the convolution theorem. We provide an example of application of image processing using the $C-$ functions and the convolutions for spatial filtering of the treated image.
\end{abstract}

\maketitle

\section{Introduction}

The development of information technologies has inspired also the development of the information compression, the most famous part of which is the image and video compression. The compression is based on the information structure in order to optimize compression speed, compression rate and the possible losses of information during the compression. Development of the theory of orbit functions opens a space for their use in the processing of the information sampled on grids in simplexes and polyhedra in $n-$dimensional space. These functions can be used for decomposition of any discrete values ​​on the grids to orthogonal series. The density of grid points is controlled by a suitable choice of parameter.  Moreover, we can glue together more simplexes and study the information carried in the grid in this ensemble. In this paper, we focus on the simplest non-trivial case of utilization of orbit functions in two dimensions. It corresponds to a two-dimensional digital image processing. In comparison with the most widespread method for image processing - Fourier analysis, i.e., the decomposition into exponential series in two perpendicular directions, we decompose discrete functions on points of the grid in a number of orbit functions without the division into several directions. Our approach is a generalization of discrete Fourier and cosine transform.

In this paper, we summarize the properties of $C -$ and $E -$orbit functions connected with Weyl groups of simple Lie algebras $A_2, C_2$ and $G_2$. These functions are a generalization of the classical cosine and exponential function and they act in fundamental domains of the Lie algebras. In these domains we introduce a discrete grid on which it is possible to define discrete $C-$ and $E-$ orbit transform. For an illustrative example of analysis and image processing we split a square image into two triangles and we effectuate ​​corresponding $C-$orbit transform.

The paper is organized as follows. Section 2 summarize some known facts about the spatial filtering using a convolution. In Section 3 we remind basic notations from the theory of Weyl group orbit functions. In particular, we describe the discrete transforms based on finite families of orbit functions in Subsection 3.3. In section 4 we define $C-$ and $E-$orbit convolution and we formulate the orbit convolutions theorems. Finally, in Subsection 4.2 we provide examples of image processing using $C-$orbit functions. We include two appendices with technical details for the orbit transforms.

\section{Spatial filtering}

A variety of filters play an important role in image processing, in image improving and in detail recognition. For example, the spatial filtering uses convolution of functions which is performed via Fourier transform as a multiplication of the Fourier images. Fourier analysis is based on the decomposition of brightness values ​​in each digitized image points along the rows and columns into Fourier series. The Fourier transform is then processed. The inverse discrete Fourier transform shows processing of digital images. This way we can highlight some features of the image -- remove the noise or enhance blur edges. The whole process is described in several papers, for an overview see for example~\cite{GoWo,GrKn}. For image compression JPEG the discrete cosine transforms are used. They are of four types and the convolution via multiplications in these cases is more complicated, it combines cosine and sine discrete transform except the discrete cosine transform of type II. 
The simplest filtering technique is the averaging the light intensities at points. Intensity of each new pixel is the mean value of the intensities of the 8 neighboring pixels and the pixel itself in the original image. Other filters use the intensities ​​of neighboring pixels multiplied by different relative weights and the pixel is assigned by a mean value of 9 intensities. Other filters take into account a number of other surrounding pixels, 25 pixels together with the center. Intensities in 9 or 25 pixel can be expressed as $3 \times 3$ or $5 \times 5$ matrix. Averaging over neighboring pixels is mathematically expressed by the convolution of the original intensity matrix with $3 \times 3$ or $5 \times 5$ matrix, so-called convolution kernel. The elements of this matrix are the weights assigned to the corresponding pixel in the area according to the desired filter type. For the treatment of pixel intensities on the edge we need to extend a line above and below the picture and a column on the left and the right in the $3 \times 3$ matrix case. In the case of $5 \times 5$ matrix we need to add to each side two columns and two rows.

Filters mentioned above are called linear spatial filters. Their application to a digital image creates a new image using a linear combination of brightness values ​​in the surrounding pixels. The intensities of the digital image in each pixel are defined by the matrix $f (m, n)$. If we want to apply a filter comprising eight neighboring pixels with different weights, we construct the $3\times 3$ weights matrix

$$\begin{pmatrix}
a_{-1-1}&a_{-10}&a_{-11}\\a_{0-1}&a_{00}&a_{01}\\a_{1-1}&a_{10}&a_{11}
\end{pmatrix}.$$
New digital image has the intensity in each pixel given by a matrix $F (m, n)$ and their values are 
$$\begin{aligned} F(m,n)&= a_{-1-1}f(m-1,n-1) +a_{-10}f(m-1,n)+a_{-11}f(m-1,n+1)\\ & +a_{0-1}f(m,n-1)+a_{00}f(m,n)+a_{01}f(m,n+1)\\ &
+a_{1-1}f(m+1,n-1)+a_{10}f(m+1,n)+a_{11}f(m+1,n+1).\\\end{aligned}
$$
This corresponds to the sum of all the values of the $3\times 3$ matrix we get as a pointwise multiplication of the filter $3\times 3$ matrix cut around the filtered pixel. Mathematically, it is a discrete convolution

$$F(m,n)=\sum_{i,j=-1}^1f(m+i,n+j)a_{ij}.$$
For defining the orbit convolutions we proceed in a similar way as for the discrete cosine transform DCT II, where for two functions $f$ and $g$ it is defined 

$$ (f*g)(x)= \frac{1}{\sqrt{ 2 \pi}} \int_0^\infty f(y)( g(|x-y|) + g(x+y))dy  $$
and for cosine transform $F_c$ the following relation holds~\cite{Kakichev}

$$F_c (f*g)(x) = (F_c f) (x) (F_c g)(x).$$

\section{Weyl group orbit functions}

\subsection{Weyl groups and affine Weyl groups}
We consider the simple Lie algebras of rank two, namely $A_2,C_2$ and $G_2$. Each of them is described by its set of simple roots $\Delta=\lb\a_1,\a_2\rb\in\R^2$. In the case of $A_2$, the roots are of the same length, for $C_2$ and $G_2$ we distinguish so-called short root and long root. We use the standard normalization $\l\a,\a\r=2$ for the long roots. Coroots are defined as $\a^\vee=2\a/\sca{\a}{\a}$. Moreover, we define the weights $\o_i$ and coweights $\ov_i$, which are dual to root and coroots in the sense 
$\sca{\a^\vee_i}{\o_j}=\sca{\a_i}{\o^\vee_j}=\delta_{ij}$. The weight lattice $P$ is defined as all integer combinations of weights.

We denote the reflections with respect to the hyperplanes orthogonal to the simple roots by $r_1$ and $r_2$, i.e., $r_{i}x=x-\langle \a_i,x\rangle\a_i^\vee. $
They generate a Weyl group corresponding to each Lie algebra. The action of $W$ on the set of simple roots gives a root system $W(\Delta)$ in $\R^2$. It contains a unique highest root $\xi=m_1\a_1 +m_2 \a_2$, where the coefficients $m_{1,2}$ are called the marks.  
Analogously, a root system $W(\Delta^\vee)$ is obtained from the action of $W$ on the set of coroots, its highest root is denoted by $\eta=m^\vee_1\a^\vee_1 +m^\vee_2 \a^\vee_2$, the coefficients are called the dual marks.

\begin{figure}[ht!]
  \centering
  \includegraphics[width=0.9\textwidth]{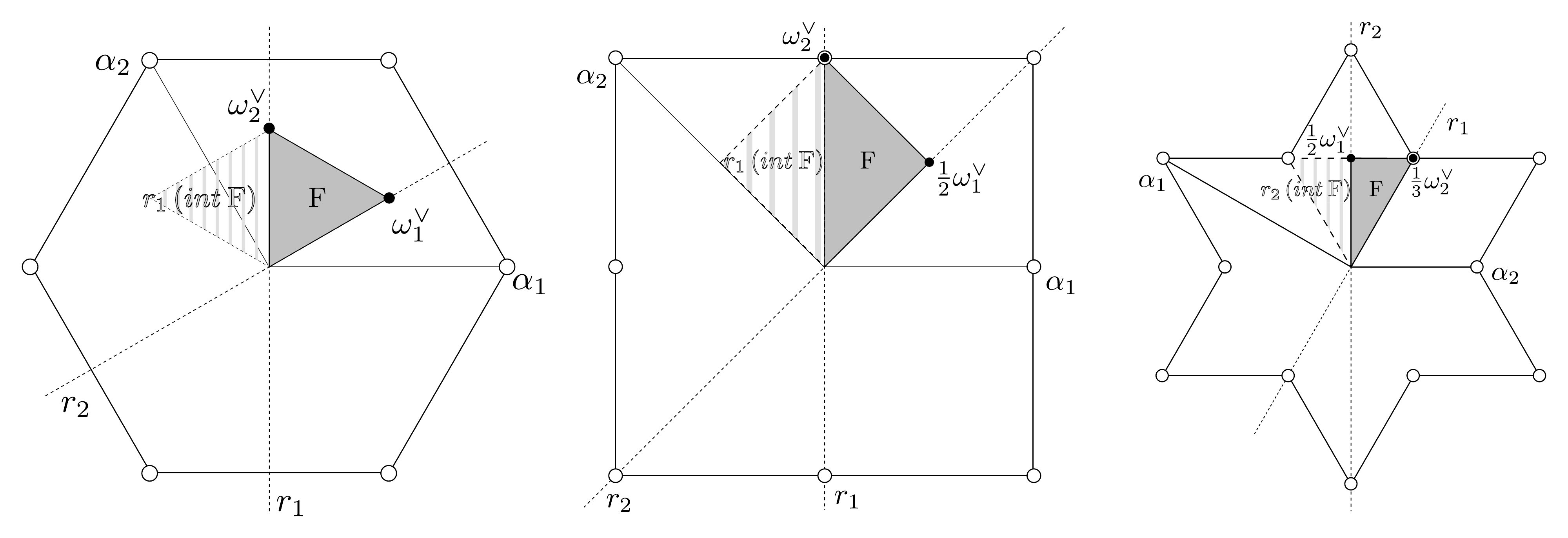}
  \caption{Root system of $A_2,C_2$ and $G_2$. Dots denotes the roots, dashed lines the hyperplanes orthogonal to the simple roots and the gray triangle is the fundamental domain of the corresponding affine Weyl group.}\label{fig1}
\end{figure}

Let $r_\xi$ denote the reflection with respect to the hyperplane orthogonal to $\xi$ and we define $r_0$ by
$$
r_0 x=r_\xi x + \frac{2\xi}{\sca{\xi}{\xi}}\,,\quad
r_{\xi}x=x-\frac{2\sca{x}{\xi} }{\sca{\xi}{\xi}}\xi\,,\quad x\in\R^2.$$
The affine Weyl group $\affW$ is generated by $\lb r_0, r_1, r_2 \rb$. Its fundamental domain is a connected subset of $\R^2$ such that it contains exactly one point of each affine Weyl group orbit. It can be chosen~\cite{HrPa01} as the convex hull of the points $\lb 0, \frac{\o^{\vee}_1}{m_1},\frac{\o^{\vee}_2}{m_2}\rb$. The root systems and the fundamental domains of affine Weyl group of $A_2,C_2$ and $G_2$ are depicted in Figure~\ref{fig1}.

The even Weyl group $W^e$ is defined as $W^e=\set{w\in W}{\det (w)=1}$. Its fundamental domain is $F^e=F\cup r_i(\inter F)$, where $r_i$ is a simple reflection and $\inter F$ denotes the interior of $F$~\cite{HrPa02}. Corresponding dual even affine Weyl group is denoted $\hat{W}^{\mathrm{aff}}_e$ and its fundamental domain is given by $F^{e\vee}=F^\vee\cup r_i\inter( F^\vee)$.

\subsection{Weyl group orbit functions}
Three families of Weyl group orbit functions, so-called $C-,S-$ and $E-$functions, are defined in the context of any Weyl group. Their complete description can be found in the series of papers~\cite{KlPa01,KlPa02,KlPa03} The family of $C-$functions is defined as follows: For every $x\in\R^2$ and $\lambda\in P$ we have
\begin{equation*}
  \Phi_\lambda (x)=\sum_{w\in W}e^{2\pi i \sca{w(\lambda)}{x}}.
\end{equation*}
The functions are invariant with respect to the affine Weyl group, therefore, we can consider $x\in F$ only. 

The family of $S-$functions is defined for every $x\in\R^2$ and $\lambda\in P$ as
\begin{equation*}
  \phi_\lambda (x)=\sum_{w\in W}\det(w)e^{2\pi i \sca{w(\lambda)}{x}}.
\end{equation*}
They are antiinvariant with respect to $\affW$, moreover, they vanish on the boundary of the fundamental domain. We can consider $x\in \inter F$.

Finally, the $E-$orbit functions are defined for every $x\in\R^2$ and $\lambda\in P$ as
\begin{equation*}
  \Xi_\lambda (x)=\sum_{w\in W^e}e^{2\pi i \sca{w(\lambda)}{x}}.
\end{equation*}
They are invariant with respect to the even affine Weyl group, we restrict them on $F^e$.

For Weyl groups with two different lengths of root in their root system other families of orbit functions can be defined. For more details see~\cite{HHP,MMP}.
In this paper we consider convolution based on the $C-$ and $E-$functions, $S-$function convolution does not differ significantly from the $C-$function case. 


\subsection{Discrete orthogonality and orbit transform}
The method of discretization of orbit functions was described in detail in the papers~\cite{HrPa01, HrPa02}. The general idea is the following: In the fundamental domain we define a finite grid of points $F_M$, where $M$ is an integer of our choice which allows us to control the density of the grid. A discrete scalar product of functions is then defined using this points. We describe a finite family of orbit functions which are pairwise orthogonal with respect to this scalar product by defining a grid of parameters labeling the functions. Finally, we give the explicit orthogonality relations. Appendix~\ref{ap1} summarize details about the choice of the grids.

We consider a space of discrete functions sampled on the points of $F_M$ with a scalar product defined for each pair of functions $f,g$ as
\begin{equation}\label{sca}
  \sca{f}{g}_{F_M}=\sum_{x\in F_M}\ep(x)f(x)\overline{g(x)}\,.
\end{equation}
The weight function $\ep(x)$ is given by the order of the Weyl orbit of $x$, $\ep(x)=\frac{|W|}{|\stab_W(x)|}$. 
The set of parameters $\Lambda_M$ gives us a finite family of orbit functions which are pairwise orthogonal with respect to the scalar product~\eqref{sca}.

For every $\la,\la'\in\Lambda_M$ it holds that
\begin{equation}\label{OG}
  \sca{\Phi_\la}{\Phi_{\la'}}_{F_M}=c|W|M^2h_\la^\vee \delta_{\la\la'}, \end{equation}
where the coefficient $h_\la^\vee$ is the order of the stabilizer of $\la$, $c$ is determinant of the Cartan matrix of the corresponding Weyl group and $|W|$ is its order. The values of $|W|$, $c$, $\ep(x)$ and $h^\vee_\la$ are listed in Appendix~\ref{ap2}.

The discrete orthogonality allows us to perform a Fourier like transform, called $C-$orbit transform. We consider a function $f$ sampled on the points of $F_M$. We can interpolate it by a sum of $C-$functions
\begin{equation}\label{transform}
  I_M(x)=\sum_{\la\in\Lambda_M}F_\la\Phi_\la(x),
\end{equation}
where we require $f(x)=I_M(x)$ for every $x\in F_M$. Therefore, the coefficients $F_\la$ are equal to
\begin{equation}\label{coef}
\begin{split}
  F_\la&=\frac{\sca{f}{\Phi_\la}_{F_M}}{\sca{\Phi_\la}{\Phi_\la}_{F_M}}\\
  &=\frac{1}{c|W|M^2h_\la^\vee}\sum_{x\in F_M} \ep(x)f(x)\overline{\Phi_\la(x)}.
\end{split}
\end{equation}

In the case of $E-$orbit functions we consider the grids $F^e_M$ and $\Lambda^e_M$. The scalar product is defined as
\begin{equation}\label{sca2}
  \sca{f}{g}_{F^e_M}=\sum_{x\in F^e_M}\ep^e(x)f(x)\overline{g(x)}\,.
\end{equation}
The weight function $\ep^e(x)$ is given by the order of the even Weyl orbit of $x$, $\ep^e(x)=\frac{|W^e|}{|\stab_{W^e}(x)|}$. 

For every $\la,\la'\in\Lambda^e_M$ it holds that
\begin{equation}\label{OG2}
  \sca{\Xi_\la}{\Xi_{\la'}}_{F^e_M}=c|W^e|M^2h_\la^{e\vee} \delta_{\la\la'}, \end{equation}
where the coefficient $h_\la^{e\vee}$ is the order of the stabilizer of $\la$ and $|W^e|$ is the order of the even Weyl group.. The values of $|W^e|$, $\ep^e(x)$ and $h^{e\vee}_\la$ are listed in Appendix~\ref{ap2}.

The $E-$orbit transform is provided as follows. We consider a function $f$ sampled on the points of $F^e_M$. We can interpolate it by a sum of $E-$functions
\begin{equation}\label{transform2}
  I_M(x)=\sum_{\la\in\Lambda^e_M}F_\la\Xi_\la(x),
\end{equation}
where we require $f(x)=I^e_M(x)$ for every $x\in F^e_M$. Therefore, the coefficients $F_\la$ are equal to

\begin{equation}\label{coef2}
  F_\la=\frac{\sca{f}{\Xi_\la}_{F^e_M}}{\sca{\Xi_\la}{\Xi_\la}_{F^e_M}}=\frac{1}{c|W^e|M^2h_\la^\vee}\sum_{x\in F^e_M} \ep(x)f(x)\overline{\Xi_\la(x)}.
\end{equation}

\section{Orbit convolution}

\subsection{Orbit convolution theorem}
The main aim of this work is to define a discrete orbit functions convolution, i.e., a mapping of two functions sampled on $F_M$ which respects a relation analogous to the classical convolution theorem. Such definition comes naturally from the orbit functions discretization theory.

The $C$-orbit convolution is for every pair of discrete functions $f,g$ and $u\in F_M$ defined as
\begin{equation}\label{convolution}
  (f\ast g)(u) \ldef \sum_{x\in F_M}\ep(x) \sum_{w \in W} f(x)g(u-w(x)).
\end{equation}
Such a convolution is well defined, the shifts in the convolution kernel $g$ respect the symmetry of the Weyl group of $A_2$. We can write the $C-$orbit convolution theorem.

\begin{theorem}
  Let $f,g$ be any functions defined on the points of $F_M$ and $u\in F_M$. Then
  \begin{equation}\label{convolutionTheorem}
    (f\ast g)(u)=\sum_{\la\in\Lambda_M} c|W| M^2 h_\la^\vee F_\la G_\la \Phi_\la(u),
  \end{equation}
  where $F_\la$ and $G_\la$ are the $C-$orbit transforms of $f$ and $g$ given by~\eqref{transform}.
\end{theorem}

Its proof is straightforward, it uses the relations~\eqref{coef} and the following formula for the product of an orbit function with the complex conjugate of an orbit function with the same label but different argument:
\begin{equation}\label{product}
  \Phi_{\lambda}(x)\overline{\Phi_{\lambda}(y)}=\sum_{w\in W}\Phi_{\la}(x-w(y)).
\end{equation}

Analogously, the $E$-orbit convolution is defined for discrete functions $f,g$ sampled on $F^e_M$ and $u\in F^e_M$ as
\begin{equation}
  (f\ast g)(u) \ldef \sum_{x\in F^e_M}\ep(x) \sum_{w \in W^e} f(x)g(u-w(x)).
\end{equation}
The $E-$orbit convolution theorem is then the following.

\begin{theorem}
  Let $f,g$ be any functions defined on the points of $F^e_M$ and $u\in F^e_M$. Then
  \begin{equation}\label{convolution2}
    (f\ast g)(u)=\sum_{\la\in\Lambda^e_M} c|W| M^2 h_\la^\vee F_\la G_\la \Xi_\la(u),
  \end{equation}
  where $F_\la$ and $G_\la$ are the $E-$orbit transforms of $f$ and $g$ given by~\eqref{transform2}.
\end{theorem}

\subsection{Examples of image filtering}
For the purpose of demonstrating the differences between the orbit convolution and convolution on $\R^2$ we take an artificial image of a hexagon. Three of spatial filters are presented: a mean filter, often used for image denoising; a sharpen filter which is useful for contrast enhancing; and a simple edge detecting filter which suppresses the monotonic (in the sense of pixel brightness) parts of an image.

In $\R^2$ these filters are described by matrices:
\begin{gather*}
		h_{\scriptscriptstyle \text{mean}} = \frac{1}{9} \begin{pmatrix}	1 & 1 & 1 \\	1 & 1 & 1 \\	1 & 1 & 1 \end{pmatrix}, \quad
		h_{\scriptscriptstyle \text{sharpen}} = \begin{pmatrix}	0 & -1 & 0 \\	-1 & 5 & -1	\\	0 & -1 & 0 \end{pmatrix}, \\
		h_{\scriptscriptstyle \text{edge}} = \begin{pmatrix}	0 & 0 & -1 \\	-1 & 3 & 0 \\	0 & 0 & -1 \end{pmatrix}.
\end{gather*}

The filters are constructed to be as similar to the filters used for orbit convolution as possible. There are some restrictions for the orbit convolution coming from its definition, the most significant is the summation over all reflections of the convolution kernel. This property is unpleasant, since it does not give us the possibility to apply changes in a single direction, i.e., detecting only horizontal edges. For this reason we cannot use all convolution kernels we can use for image filtering in $\R^2$.

When developing a spatial filter for orbit convolution from kernel for filtering in $\R^2$ we have to take the formula \eqref{convolution} into account. Many filters are supposed to preserve the average value of brightness in the image. In the frequency domain the related value is situated in the point $\left( 0, 0 \right)$. The normalization of the filter is done by dividing the weighted sum of kernel points by coefficients $\varepsilon(x)$. There is also a second level of normalization, arising from the summation over all Weyl reflections of a point, the filter is divided by the number of reflections. Some filters, mostly the ones based on differences, have the weighted sum equal to zero, thus not requiring any normalization.

There are two major restrictions for the orbit convolution kernels: the reflection of the kernel, which disables filtering in a single direction, and the placement of the kernel center. For the convolution on $\R^2$ the kernel center is located in the middle point of the kernel, for orbit convolution the center is in the point $\left( 0, 0 \right)$. This brings further restriction, the filter cannot count with all neighboring points.

Filters for orbit convolution are defined in the following way:

\begin{gather*}
	h_{\scriptscriptstyle \text{mean}} = \frac{1}{3} \begin{pmatrix}	\begin{matrix} 1 \\ 1 \end{matrix} & 1 \end{pmatrix} , \quad
	h_{\scriptscriptstyle \text{sharpen}} =  \begin{pmatrix} \begin{matrix} 0 \\ 5 \end{matrix} & -1 \end{pmatrix} , \\
	h_{\scriptscriptstyle \text{edge}} =  \begin{pmatrix} \begin{matrix} 0 \\ 3 \end{matrix} & -1 \end{pmatrix} .
\end{gather*}

\begin{figure}[ht!]
	\centering
	\includegraphics[width=.25\textwidth]{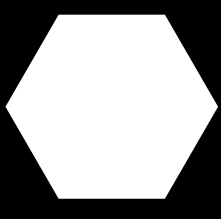}
	\caption{Original image, used for filtering}\label{fig2}
\end{figure}

\begin{figure}[ht!]
	\centering
	\includegraphics[width=.5\textwidth]{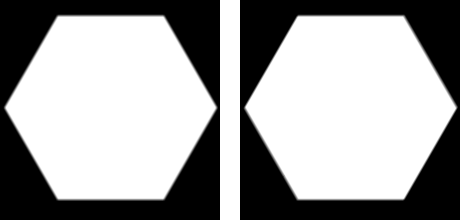}
	\caption{Image filtered with blurring filter, via $\R^2$ convolution (left) and orbit convolution (right)}\label{fig3}
\end{figure}

\begin{figure}[ht!]
	\centering
	\includegraphics[width=.5\textwidth]{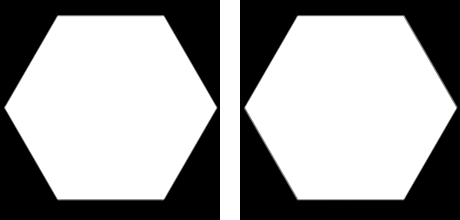}
	\caption{Sharpening previously blurred image, $\R^2$ (left) and orbit (right) convolution}\label{fig4}
\end{figure}

\begin{figure}[ht!]
	\centering
	\includegraphics[width=.5\textwidth]{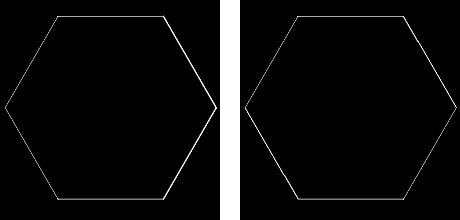}
	\caption{Edge detection in the original image, on the left with $\R^2$ convolution, on the right with orbit one}\label{fig5}
\end{figure}

For the orbit convolution demonstration we used the hexagon image, see Figure~\ref{fig2}, and filtered it via convolution on $\R^2$ and via $C-$orbit convolution on $A_2$ group to have a comparison for similar filters for both methods. The results are depicted on Figures~\ref{fig3},~\ref{fig4} and~\ref{fig5}.

The differences between the convolution on $\R^2$ and orbit convolution via $C$-orbit transform on $A_2$ group are very little. One of the reasons is the inequality of convolution kernels for both types of convolution.

\section{Concluding remarks}

\begin{enumerate}
 \item In the case of $C_2$ and $G_2$ orbit functions there are 7 more families of orbit function defined, the orbit convolution theorem can be formulated for each of them. This gives us bigger choice of the shape of the fundamental domain suitable for the image.
 \item The method described here can be generalized to Weyl group of any rank. Therefore, it can be used for more general problems than the image processing.
 \item The orbit convolution takes an advantage from the symmetry of the underlying Weyl group. On the other hand, as there is no fast algorithm yet, the computation takes more time than standard Fourier or cosine transform. One of our future projects is finding such a fast algorithm. 
\end{enumerate}

\section*{Acknowledgment}
This work is supported by the European Union under the project “Support of inter-sectoral mobility and quality enhancement of research teams at Czech Technical University in Prague” CZ.1.07/2.3.00/30.0034.

\section{Appendix}

\subsection{Grids $F_M$ and $\Lambda_M$}\label{ap1}
In this section we describe in detail the grids of points and grids of parameters used in the discretization of orbit functions~\cite{HrPa01, HrPa02}.

We consider four lattices in $\R^2$. The root lattice $Q=\Z\a_1+\Z\a_2$; the coroot lattice $Q^\vee=\Z\a_1^\vee+\Z\a_2^\vee$ and their duals $P=\Z\o_1+\Z\o_2$ and $P^\vee=\Z\o^\vee_1+\Z\o^\vee_2$ which are called the weight lattice and coweight lattice respectively. 

Two finite lattice grids depending on an integer parameter $M$ are defined as follows: We consider the $W-$invariant group $\frac{1}{M} P^\vee / Q^\vee$ and we define the set of points $F_M$ as such cosets from $\frac{1}{M} P^\vee / Q^\vee$ which have a representative in $F$. It can be written as

$$F_M=\set{\frac{s_1}{M}\ov_1+\frac{s_2}{M}\ov_2}{s_0,s_1,s_2\in \Z^{\ge0},\ s_0+s_1m_1+s_2m_2=M}.$$

The explicit formulas are then obtained by using the marks $m_1$ and $m_2$ of the concrete group. Namely, the marks are $(1,1)$ for $A_2$, $(2,1)$ for $C_2$ and $(2,3)$ for $G_2$.

\begin{equation}\label{FM}\begin{aligned}
  A_2:\quad F_M&=\set{ \frac{s_1}{M}\o^\vee_1 + \frac{s_2}{M}\o^\vee_2}{s_0, s_1, s_2 \in \Z^{\ge0}, s_0 + s_1 + s_2 = M},\\
  C_2:\quad F_M&=\set{ \frac{s_1}{M}\o^\vee_1 + \frac{s_2}{M}\o^\vee_2}{s_0, s_1, s_2 \in \Z^{\ge0}, s_0 + 2s_1 + s_2 = M},\\
  G_2:\quad F_M&=\set{ \frac{s_1}{M}\o^\vee_1 + \frac{s_2}{M}\o^\vee_2}{s_0, s_1, s_2 \in \Z^{\ge0}, s_0 + 2s_1 + 3s_2 = M},\\
\end{aligned} \end{equation}

For the grid of parameters we take the $W-$invariant group $\Lambda_M=P/MQ $ and we consider its cosets with a representative element in $MF^{\vee}$. Explicitly,

$$\Lambda_M=\set{t_1\o_1+t_2\o_2}{t_0,t_1,t_2\in \Z^{\ge0},t_0+t_1m^\vee_1+t_2m^\vee_2=M},$$

where the duals marks $m_1^\vee$ and $m_2^\vee$ are $(1,1)$ for $A_2$, $(1,2)$ for $C_2$ and $(3,2)$ for $G_2$.

\begin{equation}\label{LM}\begin{aligned}
  A_2:\quad\Lambda_{M} &= \set{t_1\o_1+t_2\o_2}{t_0,t_1,t_2\in\Z^{\ge0},t_0+t_1+t_2=M},\\
  C_2:\quad\Lambda_{M} &= \set{t_1\o_1+t_2\o_2}{t_0,t_1,t_2\in\Z^{\ge0},t_0+t_1+2t_2=M},\\
  G_2:\quad\Lambda_{M} &= \set{t_1\o_1+t_2\o_2}{t_0,t_1,t_2\in\Z^{\ge0},t_0+3t_1+2t_2=M}.\\
\end{aligned}\end{equation}

The grids for the $E-$transform are defined analogously, 
$$\begin{aligned}F^e_M&=\frac{1}{M} P^\vee / Q^\vee\cap F^e,\\
   \Lambda^e_M&=P/MQ\cap MF^{e\vee}=\Lambda_M\cup r_i(\inter \Lambda_M),\\
  \end{aligned}$$

\noindent where $F^e=F\cup r_i(\inter F)$ and $F^{e\vee} = F^\vee\cup r_i(\inter F^\vee)$ for a simple reflection $r_i$.
  
\subsection{Values of $|W|,c,h_\lambda^\vee$ and $\varepsilon$}\label{ap2}
We summarize values of all the constants and functions needed in formulas~\eqref{OG},~\eqref{transform},~\eqref{OG2},~\eqref{transform2}.

The orders of the corresponding Weyl groups and even Weyl groups are:
\begin{equation}\label{orders}
|W|=\begin{cases}
     6, \text{ for }A_2,\\
     8, \text{ for }C_2,\\
     12, \text{ for }G_2,\\
    \end{cases}\quad
|W^e|=\begin{cases}
     3, \text{ for }A_2,\\
     4, \text{ for }C_2,\\
     6, \text{ for }G_2,\\
    \end{cases}
\end{equation}

The determinants of the corresponding Cartan matrix are:
\begin{equation}\label{determinant}
c=\begin{cases}
     3, \text{ for }A_2,\\
     2, \text{ for }C_2,\\
     1, \text{ for }G_2.\\
    \end{cases}
\end{equation}

The values of $\ep(x)$ and $h_\lambda^\vee$ are listed in Tables~\ref{tab1} and~\ref{tab2}.

\begin{table}[ht!]
\centering
\begin{tabular}{c c|c}
  $A_2$ & $x\in F_M$ & $\ep(x)$  \\ 
  \hline
  &$[s_0,s_1,s_2]$ & $6$  \\
  &$[s_0,s_1,0]$ & $3$  \\
  &$[s_0,0,s_2]$ & $3$  \\
  &$[0,s_1,s_2]$ & $3$  \\
  &$[0,0,s_2]$ & $1$  \\
  &$[0,s_1,0]$ & $1$  \\
  &$[s_0,0,0]$ & $1$  \\
\end{tabular}\hspace{8pt}
\begin{tabular}{c c|c}
  $C_2$&$x\in F_M$ & $\ep(x)$  \\ 
  \hline
  &$[s_0,s_1,s_2]$ & $8$  \\
  &$[s_0,s_1,0]$ & $4$  \\
  &$[s_0,0,s_2]$ & $4$  \\
  &$[0,s_1,s_2]$ & $4$  \\
  &$[0,0,s_2]$ & $1$  \\
  &$[0,s_1,0]$ & $2$  \\
  &$[s_0,0,0]$ & $1$  \\
\end{tabular}\hspace{8pt}
\begin{tabular}{c c|c}
  $G_2$&$x\in F_M$ & $\ep(x)$  \\ 
  \hline
  &$[s_0,s_1,s_2]$ & $12$  \\
  &$[s_0,s_1,0]$ & $6$  \\
  &$[s_0,0,s_2]$ & $6$  \\
  &$[0,s_1,s_2]$ & $6$  \\
  &$[0,0,s_2]$ & $2$  \\
  &$[0,s_1,0]$ & $3$  \\
  &$[s_0,0,0]$ & $1$  \\
\end{tabular}\\
\medskip
\caption{The coefficients $\ep(x)$ of $A_2$, $C_2$ and $G_2$. The variables $s_i$, $i=0,1,2$, are nonnegative integers and have the same meaning as in~\eqref{FM}.}\label{tab1}
\end{table}

\begin{table}[ht!]
\centering
\begin{tabular}{c c|c}
  $A_2$&$\la\in \Lambda_M$  & $h_\lambda ^{\vee}$ \\ 
  \hline
  &$[t_0,t_1,t_2]$ & $1$  \\
  &$[t_0,t_1,0]$ & $2$  \\
  &$[t_0,0,t_2]$ & $2$  \\
  &$[0,t_1,t_2]$ & $2$  \\
  &$[0,0,t_2]$ & $6$  \\
  &$[0,t_1,0]$ & $6$  \\
  &$[t_0,0,0]$ & $6$  \\
\end{tabular}\hspace{8pt}
\begin{tabular}{c c|c}
$C_2$&$\la\in \Lambda_M$  & $h_\lambda ^{\vee}$ \\ 
  \hline
  &$[t_0,t_1,t_2]$ & $1$  \\
  &$[t_0,t_1,0]$ & $2$  \\
  &$[t_0,0,t_2]$ & $2$  \\
  &$[0,t_1,t_2]$ & $2$  \\
  &$[0,0,t_2]$ & $8$  \\
  &$[0,t_1,0]$ & $4$  \\
  &$[t_0,0,0]$ & $8$  \\
\end{tabular}\hspace{8pt}
\begin{tabular}{c c|c}
$G_2$&$\la\in \Lambda_M$  & $h_\lambda ^{\vee}$ \\ 
  \hline
  &$[t_0,t_1,t_2]$ & $1$  \\
  &$[t_0,t_1,0]$ & $2$  \\
  &$[t_0,0,t_2]$ & $2$  \\
  &$[0,t_1,t_2]$ & $2$  \\
  &$[0,0,t_2]$ & $12$  \\
  &$[0,t_1,0]$ & $4$  \\
  &$[t_0,0,0]$ & $6$  \\
\end{tabular}\\
\medskip
\caption{The coefficients $h_\lambda ^{\vee}$ of $A_2$, $C_2$ and $G_2$. The variables $t_i$, $i=0,1,2$, are nonnegative integers and have the same meaning as in~\eqref{LM}.}\label{tab2}
\end{table}

Let $x$ be in $F^e$. For $x\in r_i(\inter F)$ it holds that $\ep^e(x)=|W^e|$. The values of $\ep^e(x)$ for $x\in F$ are listed in Table~\ref{tab3}.

\begin{table}[ht!]
\centering
\begin{tabular}{c c|c}
  $A_2$ & $x\in F_M$ & $\ep^e(x)$  \\ 
  \hline
  &$[s_0,s_1,s_2]$ & $3$  \\
  &$[s_0,s_1,0]$ & $3$  \\
  &$[s_0,0,s_2]$ & $3$  \\
  &$[0,s_1,s_2]$ & $3$  \\
  &$[0,0,s_2]$ & $1$  \\
  &$[0,s_1,0]$ & $1$  \\
  &$[s_0,0,0]$ & $1$  \\
\end{tabular}\hspace{8pt}
\begin{tabular}{c c|c}
  $C_2$&$x\in F_M$ & $\ep^e(x)$  \\ 
  \hline
  &$[s_0,s_1,s_2]$ & $4$  \\
  &$[s_0,s_1,0]$ & $4$  \\
  &$[s_0,0,s_2]$ & $4$  \\
  &$[0,s_1,s_2]$ & $4$  \\
  &$[0,0,s_2]$ & $1$  \\
  &$[0,s_1,0]$ & $2$  \\
  &$[s_0,0,0]$ & $1$  \\
\end{tabular}\hspace{8pt}
\begin{tabular}{c c|c}
  $G_2$&$x\in F_M$ & $\ep^(x)$  \\ 
  \hline
  &$[s_0,s_1,s_2]$ & $6$  \\
  &$[s_0,s_1,0]$ & $6$  \\
  &$[s_0,0,s_2]$ & $6$  \\
  &$[0,s_1,s_2]$ & $6$  \\
  &$[0,0,s_2]$ & $2$  \\
  &$[0,s_1,0]$ & $3$  \\
  &$[s_0,0,0]$ & $1$  \\
\end{tabular}\\
\medskip
\caption{The coefficients $\ep^e(x)$ of $A_2$, $C_2$ and $G_2$. The variables $s_i$, $i=0,1,2$, are nonnegative integers and have the same meaning as in~\eqref{FM}.}\label{tab3}
\end{table}

Let $\lambda$ be in $\Lambda^e$. For $\lambda\in r_i(\inter F^\vee)$ it holds that $h_\lambda^{e\vee}$=1. The other values of $h_\lambda^{e\vee}$ are listed in Table~\ref{tab4}.

\begin{table}[ht!]
\centering
\begin{tabular}{c c|c}
  $A_2$&$\la\in \Lambda_M$  & $h_\lambda ^{e\vee}$ \\ 
  \hline
  &$[t_0,t_1,t_2]$ & $1$  \\
  &$[t_0,t_1,0]$ & $1$  \\
  &$[t_0,0,t_2]$ & $1$  \\
  &$[0,t_1,t_2]$ & $1$  \\
  &$[0,0,t_2]$ & $3$  \\
  &$[0,t_1,0]$ & $3$  \\
  &$[t_0,0,0]$ & $3$  \\
\end{tabular}\hspace{8pt}
\begin{tabular}{c c|c}
$C_2$&$\la\in \Lambda_M$  & $h_\lambda ^{e\vee}$ \\ 
  \hline
  &$[t_0,t_1,t_2]$ & $1$  \\
  &$[t_0,t_1,0]$ & $1$  \\
  &$[t_0,0,t_2]$ & $1$  \\
  &$[0,t_1,t_2]$ & $1$  \\
  &$[0,0,t_2]$ & $4$  \\
  &$[0,t_1,0]$ & $2$  \\
  &$[t_0,0,0]$ & $4$  \\
\end{tabular}\hspace{8pt}
\begin{tabular}{c c|c}
$G_2$&$\la\in \Lambda_M$  & $h_\lambda ^{e\vee}$ \\ 
  \hline
  &$[t_0,t_1,t_2]$ & $1$  \\
  &$[t_0,t_1,0]$ & $1$  \\
  &$[t_0,0,t_2]$ & $1$  \\
  &$[0,t_1,t_2]$ & $1$  \\
  &$[0,0,t_2]$ & $6$  \\
  &$[0,t_1,0]$ & $2$  \\
  &$[t_0,0,0]$ & $3$  \\
\end{tabular}\\
\medskip
\caption{The coefficients $h_\lambda ^{e\vee}$ of $A_2$, $C_2$ and $G_2$. The variables $t_i$, $i=0,1,2$, are nonnegative integers and have the same meaning as in~\eqref{LM}.}\label{tab4}
\end{table}

\end{document}